\documentclass[10pt,twocolumn,letterpaper]{article}

\usepackage{times}
\usepackage{epsfig}
\usepackage{graphicx}
\usepackage{amsmath}
\usepackage{amssymb}
\usepackage[utf8]{inputenc} 
\usepackage[T1]{fontenc}    
\usepackage{url}            
\usepackage{booktabs}       
\usepackage{amsfonts}       
\usepackage{nicefrac}  
\usepackage{microtype} 
\usepackage{bm}
\usepackage{bbm}
\usepackage{times}
\usepackage{authblk}
\usepackage{epsfig}
\usepackage{algorithm}
\usepackage{algpseudocode}
\usepackage{xcolor,colortbl}
\usepackage{graphicx}
\usepackage{color}
\usepackage{stmaryrd}

\usepackage[pagebackref=true,breaklinks=true,letterpaper=true,colorlinks,bookmarks=false]{hyperref}

\begin{document}

\title{The Missing Data Encoder: Cross-Channel Image Completion\\with Hide-And-Seek Adversarial Network}

\author[1]{Arnaud Dapogny}
\author[1, 2]{Matthieu Cord}
\author[2]{Patrick Perez}

\affil[1]{\small LIP6, Sorbonne Universit\'e, 4 Place Jussieu, Paris, France}
\affil[2]{\small Valeo.ai, Paris, France}

\date{}
\maketitle

\begin{abstract}
	Image completion is the problem of generating whole images from fragments only. It encompasses inpainting (generating a patch given its surrounding), reverse inpainting/extrapolation (generating the periphery given the central patch) as well as colorization (generating one or several channels given other ones). In this paper, we employ a deep network to perform image completion, with adversarial training as well as perceptual and completion losses, and call it the ``missing data encoder'' (MDE). We consider several configurations based on how the seed fragments are chosen. We show that training MDE for ``random extrapolation and colorization'' (MDE-REC), i.e. using random channel-independent fragments,  allows a better capture of the image semantics and geometry.
	MDE training makes use of a novel ``hide-and-seek'' adversarial loss, where the discriminator seeks the original non-masked regions, while the generator tries to hide them. We validate our models both qualitatively and quantitatively on several datasets, showing their interest for image completion, unsupervised representation learning as well as face occlusion handling.
\end{abstract}

\section{Introduction}

	In this paper, we investigate the problem of image completion, \textit{i.e.} the one of generating a complete image from RGB or single-channel parts of an original image.
	From a representation learning standpoint, learning to perform image completion amounts to encoding the underlying structures of the visual objects.
	A number of approaches have been proposed in the literature that try to learn this structure in an unsupervised fashion, in the hope that the representations learned by doing so could help other (mostly supervised) tasks, such as image classification, object detection or semantic segmentation. Indeed, for a number of these tasks, performing a supervised pre-training on a large database such as ImageNet is beneficial to the accuracy. Yet, collecting such vast amounts of data is tedious, if not impractical.
	
	\begin{figure}[ht]
		\centering
		\includegraphics[width=\linewidth]{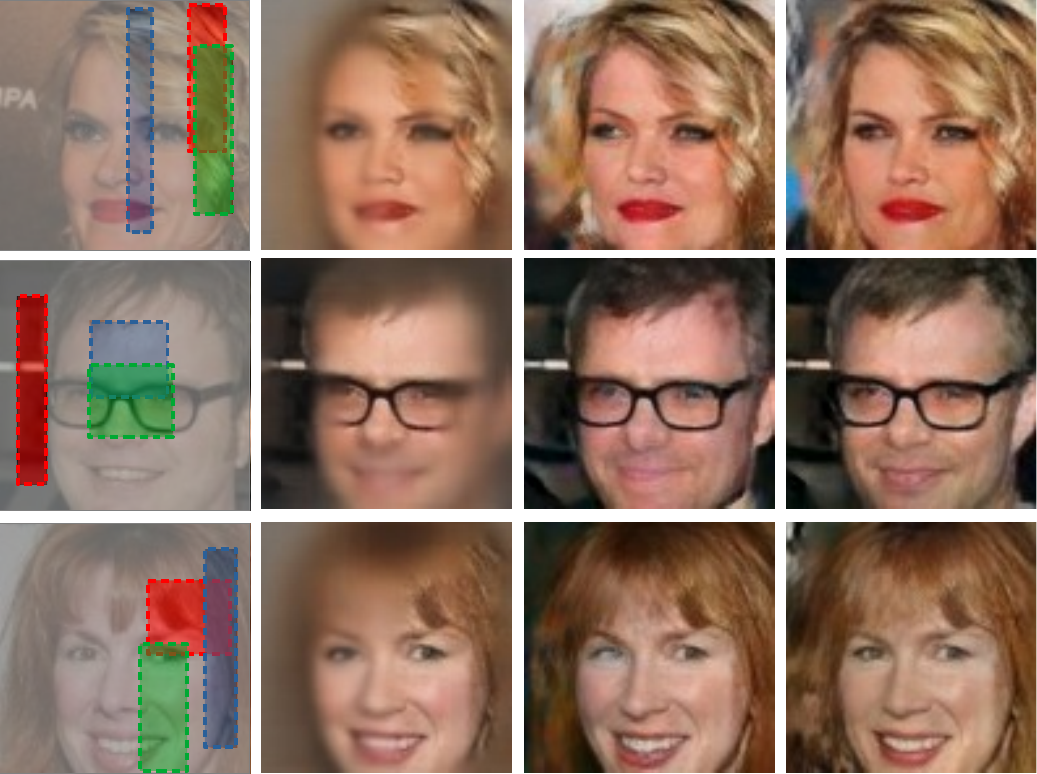}
		\caption{\textbf{Image completion from a small fragment in each color channel}. From left to right: Original image masked channel-wise; Images generated with proposed missing data encoder, trained respectively with \textit{completion}, \textit{perceptual}+\textit{adversarial} and \textit{perceptual}+\textit{adversarial}+\textit{hide-and-seek} losses (see text for details). In all cases, the image is completed using only the information within the boxes. The hide-and-seek loss ensures that there is no trace left of the generation process in the completed images.}
		\label{losses}
	\end{figure}
	
	\begin{figure*}[h!]
		\centering
		\includegraphics[width=0.93\linewidth]{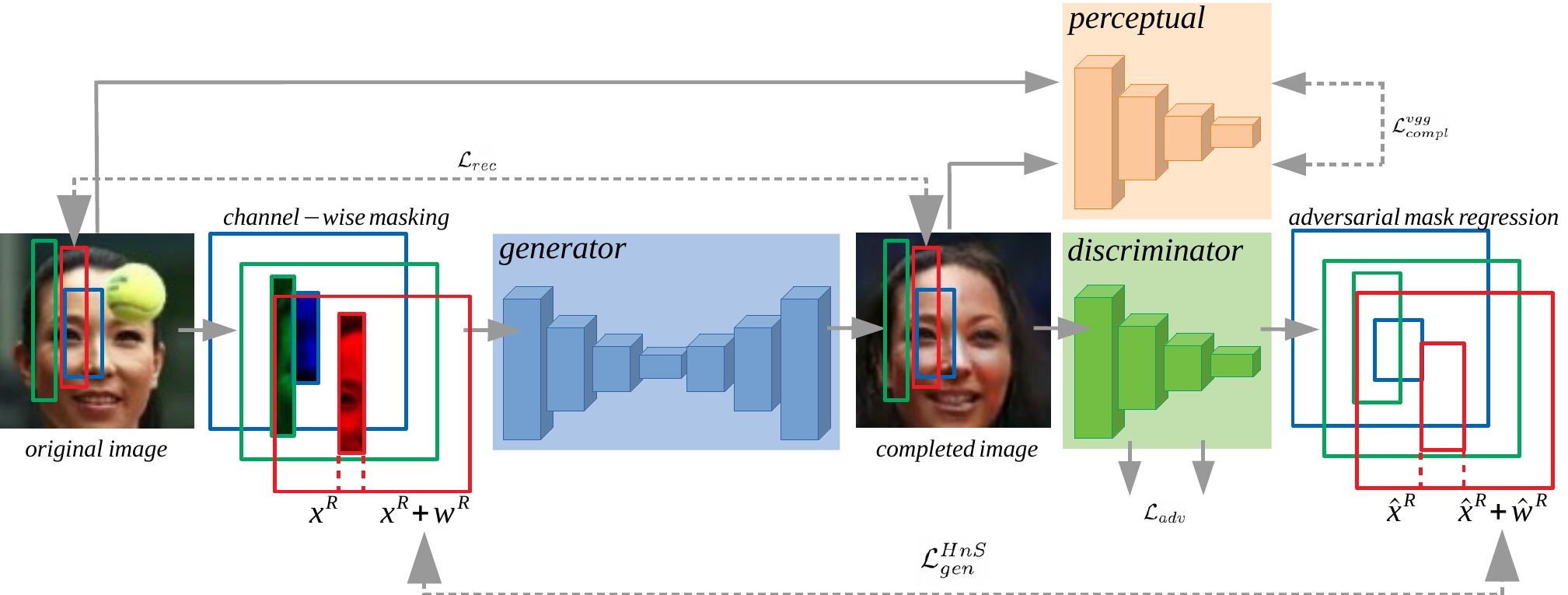}
		\caption{\textbf{Overview of the Missing Data Encoder approach}. At train time, random channel-wise masking is applied to the image, which then passes through the generator that completes it. To this end, MDE training uses perceptual, reconstruction and adversarial losses. The latter includes a novel mask regression term requiring the discriminator to ``seek'' the mask coordinates while the generator tries to``hide'' them.}
		\label{overview}
	\end{figure*}

	Following recent advances in the field of text understanding \cite{graves2013generating}, one can wonder if using the regularity of the images in an unsupervised fashion would yield such representations at virtually no cost. This echoes the ideas introduced in \cite{mathieu2015deep}, where it is theorized that a strong artificial intelligence model should build an inner representation through unsupervised learning. A general idea for doing so is to design a proxy task for pretraining. The authors in \cite{doersch2015unsupervised} proposed to predict
	the relative position of two adjacent image patches from their content.
	In the same vein, the authors in \cite{noroozi2016unsupervised} trained a network to solve jigsaw puzzles, created by shuffling a grid of patches. Intuitively, the network has to learn the structure of the objects to correctly predict the patches spatial layout and to solve the puzzle. Image colorization \cite{larsson2017colorization} has also been used as a proxy task: in \cite{zhang2017split}, the authors introduce the split-brain autoencoder, where each encoder aims at reconstructing a specific channel (\textit{e.g.} a color or a depth channel) given another one. Achieving such a completion task requires an even better capture of the visual structures by the trained network, as compared to predicting only loose spatial layout as in \cite{doersch2015unsupervised,noroozi2016unsupervised}.
	Yet, colorization is a restricted form of completion where only low frequency chrominance information needs to be inferred.
	Other proxy tasks have been proposed, such as learning motion-based segmentation in videos \cite{pathak2017learning}. Some approaches involve completing images given only a fraction of the original image. To this end, the authors of \cite{oord2016pixel} use recurrent networks to encode the spatial dependency of pixels for image completion and generation. However, the learned representations cannot be easily transferred to other tasks, as most models now involve convolutional networks. A particular case of completion is inpainting, where a central patch is reconstructed given its context, as in \cite{pathak2016context}. Similarly, Li \textit{et al.} \cite{li2017generative} propose a generative face completion method. These approaches generally rely on adversarial training \cite{goodfellow2014generative}, where a discriminator network aims at distinguishing the fake data, provided by the generator network, from the true data.
	
	Lastly, the problem of completion is related with the work in \cite{mathieu2015deep}, where the authors generate a new frame given the past frames in a video. While the setups are different, we can draw a parallel between the temporal dependency between two events, and the spatial dependency between objects in an image. For instance, a man's trajectory is predictable in the short-term as it usually varies smoothly and in relation with a context. Similarly, if we see a mug in an image we are likely to also observe a desk, or a hand.
	
	In this paper, we propose a framework for image completion using a deep neural network that we call the missing data encoder (MDE). We study several image completion scenarios with MDE: inpainting, reverse inpainting, colorization and the more general task of completing from random fragments in the different color channels -- we call it the ``random extrapolation and colorization'' (REC). The latter proves to be the best at capturing the visual semantics for subsequent use. MDE uses skip-connections to ensure that the input image regions are not altered, and is trained with a combination of completion losses, adversarial discriminative loss, perceptual loss and a novel adversarial hide-and-seek loss, as shown on Figure \ref{losses}. We demonstrate on multiple datasets that we can extrapolate high quality images from only small seed fragments, and that MDE-REC encodes semantic information as well as object geometry. The contributions of this paper are three-fold:
	
	\begin{itemize}
		\item We introduce MDE, a framework for image completion that uses a u-net-like architecture, adversarial training as well as perceptual and completion losses. We study several configurations and show that the best performing model, MDE-REC, uses a channel-wise random masking which encompasses inpainting, reverse inpainting and colorization as special cases. 
		
		\item We introduce a novel adversarial hide-and-seek loss that complements the standard adversarial objective function for image completion tasks, by specifically ensuring that there is no trace left of the generation process in the completed images.
		
		\item We thoroughly validate our model on multiple datasets, showing that MDE-REC encodes image geometry and semantics. We show several applications of MDE-REC including image generation, representation learning, and face completion under targetted occlusions.
	\end{itemize}
	
	\section{The missing data encoder}
	
	Figure \ref{overview} provides an overview of MDE-REC. As it was done in \cite{pathak2016context} for inpainting and in \cite{mathieu2015deep} for video frame prediction, we use GANs as our base architectural brick. 
	
	Given an RGB image $Z$ of size $W \times H \times 3$, we mask it by element-wise multiplication with a random binary mask $M$ of same size. As we will see in what follows, this mask can be generated in different ways.
	The generator $G$ with parameters $\theta_g$ maps the masked image $M \odot Z$ to a \textit{complete} image $G_{\theta_g}(M \odot Z)$. This new image can be decomposed as a \textit{reconstructed} region, $M \odot G_{\theta_g}(M \odot Z)$ that should closely resemble the original fragment $M \odot Z$, and a \textit{completed} one, $(1-M) \odot G_{\theta_g}(M \odot Z)$. The discriminator $D$ with parameters $\theta_d$ has to distinguish the generated images  from the real ones. Given an image training set $\{Z_i\}_{1=1}^N$ and associated masks $\{M_i\}_{1=1}^N$, this is obtained by minimizing: 
	\begin{equation}\label{eqgan}
	\footnotesize
	\mathcal{L}_{Disc}(\theta_d)=-\frac{1}{N}\sum\limits_{i=1}^N 
	\log D_{\theta_d}(Z_i) + \log[1-D_{\theta_d}(G_{\theta_g}(M_i \odot Z_i))].
	\end{equation}
	The generator has to fool the discriminator by minimizing:
	\begin{equation}\label{eqgan}
	\mathcal{L}_{Gen}(\theta_g)=-\frac{1}{N}\sum\limits_{i=1}^N \log D_{\theta_d}(G_{\theta_g}(M_{i} \odot Z_i)).
	\end{equation}
	
	In practice, optimizing solely $\mathcal{L}_{adv}(\theta_g,\theta_d)=\mathcal{L}_{Gen}(\theta_g)+\mathcal{L}_{Disc}(\theta_d)$ at train time leads to unstable behaviors. To avoid this, a classic approach \cite{pathak2016context,li2017generative} consists in adding an $L_2$ completion loss between the completed and the original regions: 
	\begin{equation}
	\mathcal{L}_{compl}(\theta_g)=\frac{1}{N}\sum\limits_{i=1}^N 
	\left\|(1-M_{i}) \odot (G_{\theta_g}(M_{i} \odot Z_i)-Z_i)\right\|_2^2.
	\end{equation}
	
	However, optimizing $\mathcal{L}_{compl}(\theta_g) + \lambda_{adv} \mathcal{L}_{adv}(\theta_g,\theta_d)$ leads to bad results, as the discriminator network quickly wins against the generator, which generates unrealistic images. Also, nothing prevents the generated image to differ from the original one on the non-masked regions.  
	
	\subsection{Preserving input information}
	
	The authors of \cite{pathak2016context} use an overlap between the inpainted region and the context, and apply a strong penalty for bad reconstructions of this region to ``guide'' training. In this vein, we add a reconstruction loss on the non-masked regions:
	\begin{equation}
	\mathcal{L}_ {rec}(\theta_g)=\frac{1}{N}\sum\limits_{i=1}^N \|M_{i} \odot (G_{\theta_g}(M_{i} \odot Z_i)-Z_i)\|_2^2.
	\end{equation}
	
	Note that such a task merely consists in autoencoding the original element: it is way easier than the task of completion and thus effectively serves as a guide for the latter task. Note that it is crucial to reconstruct the original element with high fidelity. In practice, we observe that, even if we apply a large cost to bad reconstruction of the non-masked regions, these regions are often modified. This is problematic since, in that case, the extrapolated regions do not exactly match the input information at the mask boundary. To address this problem, we use a u-net-like architecture, with skip-connections between the encoder and decoder to help preserve further the input regions.
	
	\subsection{Perceptual loss}
	
	One way to better complement the adversarial loss is to add a completion loss not directly in the image space, which results in blurry images, but in the representation space of a pretrained network such as VGG-16. As it has been pointed out \cite{johnson2016perceptual}, the first layers of a VGG network trained on large databases such as ImageNet learn filters related to image structures at different scales. Comparing images through such deep features rather than pixel-wise intensities is thus more meaningful in terms of visual structure and semantics.
	This so-called ``perceptual'' loss can be written:
	\begin{equation}
	\small
	\mathcal{L}_{compl}^{vgg}(\theta_g)=\frac{1}{N \sum\limits_{\ell=1}^L \lambda_{\ell}}\sum\limits_{\ell=1}^L \sum\limits_{i=1}^N \lambda_{\ell} 
	\left\| \phi_l( G_{\theta}(M_{i} \odot Z_i)) - \phi_l(Z_i) \right\|_2^2,
	\end{equation}
	where $\phi_{\ell}$ denotes the output of the $\ell^{\mathrm{th}}$ layer of VGG-16 and, classically, $\lambda_1=1$, $\lambda_2=0.5$, $\lambda_3=0.25$, $\lambda_4=0.125$, $\lambda_5=0.0625$ and $\lambda_{\ell}=0$ for all the fully-connected layers.
	
	\subsection{Mask generation}\label{masksgen}
	
	\begin{figure*}[htb]
		\centering
		\includegraphics[width=\linewidth]{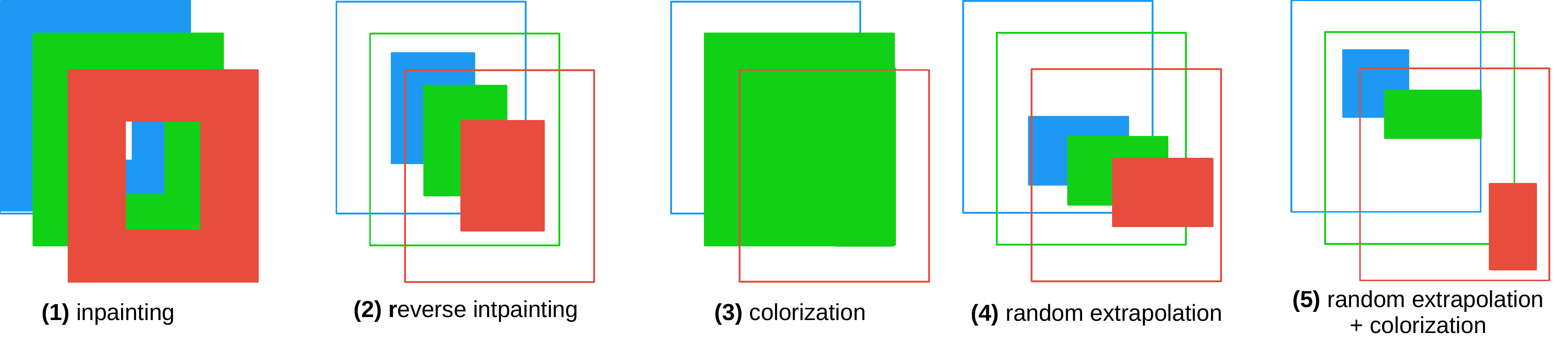}
		\vspace{-10pt}
		\caption{\textbf{Different masking methods for different image completion tasks}. (1) Inpainting (I): complete center given periphery; (2) Reverse Inpainting (RI): complete periphery given center: (3) Colorization: complete one or two color channels given the rest; (4) Random Extrapolation (RE): as RI but with a random known region; (5) Random Extrapolation and Colorization (REC): the most general task of completing image from independent random masking in the three channels.}
		\label{masks}
	\end{figure*}

	During training, for each RGB image $Z$  we generate a binary mask $M=(M^c)_{c=1}^3$ over the image channels. For each channel, $M^c$ is defined by a rectangle $R^c(S) = (x^c, y^c, w^c, h^c)$ of size $w^c\times h^c$, lower-left corner $(x^c,y^c)$ and area $SWH$, with $S\in(0,1)$ the image masking ratio hyperparameter. Figure \ref{masks} summarizes different configurations for the mask generation process. Note that except for inpainting, in all of them, the rectangle interior defines the un-masked image region, the one that the network sees.  
	The most general masking is used to perform ``random extrapolation and colorization'' (REC, Fig. \ref{masks}(5)). This task amounts to the completion of the image over the intersection of the three channel-wise masked regions and the colorization of remaining regions.
	Mask sampling is done as follows in each channel independently:
	\begin{equation}
	\begin{split}
	    & x^c \sim U(0,W-w^c),~y^c \sim U(0,H-h^c) \\
	    & h^c \sim U(SH,H),~w_c = SWH/h^c\\
	    & M^c(x,y) = \boldsymbol{1}_{[x^c<x\leq x^c+w^c]} \boldsymbol{1}_{[y^c<y\leq y^c+h^c]},
	\end{split}
	\label{eq:rec_sampling}
	\end{equation}
	where $U(a,b)$ denotes the uniform distribution over interval $[a,b]$.
	
	The other completion tasks are special cases of REC. For instance, random extrapolation (RE, Fig. \ref{masks}(4)) is the particular case where the masks are the same for all channels. For colorization (Fig. \ref{masks} (3)), the mask covers the entirety of one or two channels and nothing of remaining channels,
	Reverse Inpainting (RI, Figure \ref{masks}(2)) is obtained from RE by fixing the mask coordinates and dimensions. Finally, inpainting (I, Figure \ref{masks}(1)) is obtained from RI by switching the binary mask $M$ to $1-M$.

    The proportion of dropped pixels (\textit{i.e.} those for which all channels are missing) in the RI and RE tasks is exactly $1-S$. In the general case of REC, when boxes are different across channels, an average proportion of $(1-S)^3$ pixels is dropped and an average proportion of $1-S^3$ is corrupted (at least one channel is missing). When $S = 0.1$ as in most of our experiments, this amounts to 72.9\% (resp. 99.9\%) of dropped (resp. corrupted) pixels in average.  
    
	\subsection{Hide-and-seek loss}
	
	Despite the use of adversarial training and perceptual loss, the generator quickly learns to reconstruct the non-masked regions, which results in discontinuities on the boundaries of the masked regions. To avoid this, we design a novel adversarial mask coordinates regression loss for the discriminator, which shall estimate the locations of the original input masks by looking at the generated images (for MDE-RE and MDE-REC). Formally, we denote $r^{c}= (x^c/W, y^c/H, (x^c+w^c)/W, (y^c+h^c)/H)$ the normalized lower-left and upper-right coordinates of the ground truth box for channel $c$ and $\hat{r}^c(\theta_g,\theta_d)$ a $4$-dimensional sigmoid layer added at the end of the discriminator network (one for each channel). The adversarial mask regression loss reads: 
	\begin{equation}
	\mathcal{L}^{HnS}_{disc}(\theta_d)=\frac{1}{N}\sum\limits_{i=1}^N\sum\limits_{c=1}^C
	\|r_i^{c}-\hat{r}_i^c(\theta_g,\theta_d)\|.
	\end{equation}
	In case of a fake image, this loss makes the discriminator ``seek'' the original mask. On the other hand, the generator tries to ``hide'' it from the discriminator, \textit{e.g.} by assigning to the regressed values $\hat{r}_i^c$ the coordinates of a randomly generated box (one per epoch) $q_i^c$:
	\begin{equation}
	\mathcal{L}^{HnS}_{gen}(\theta_g)=\frac{1}{N}\sum\limits_{i=1}^N\sum\limits_{c=1}^C
	\|q_i^{c}-\hat{r}_i^c(\theta_g,\theta_d)\|.
	\end{equation}
	
	We refer to the sum of these losses as $\mathcal{L}^{HnS}(\theta_g,\theta_d)$. Note that in case of a real image, this loss is simply not used. In other words, this new game between the generator and discriminator networks ensures that there is no trace left of the generation process within the images, hence it helps reduce the artifacts caused by adversarial training. However, as pointed out in \cite{liu2018intriguing}, regressing coordinates is a hard task for convolutional networks as their structure enforces translational invariance. Thus, we also experiment with concatenating 2 channels containing $x$ and $y$-coordinates to the discriminator's inputs. We refer to this version as $\mathcal{L}^{HnS}_{coord}$. Our total loss is:
	\begin{multline}
	\mathcal{L}_{tot}(\theta_g,\theta_d) =
	\mathcal{L}_{rec}(\theta_g) + \lambda_{compl} \mathcal{L}_{compl}^{vgg}(\theta_g) \\
	+ \lambda_{adv}\mathcal{L}_{adv}(\theta_g,\theta_d)+\lambda_{HnS}\mathcal{L}^{HnS}_{coord}(\theta_g,\theta_d).
	\end{multline}
	
	\subsection{Implementation details}\label{archi}
	
\begin{figure}[htb]
		\centering
		\includegraphics[width=\linewidth]{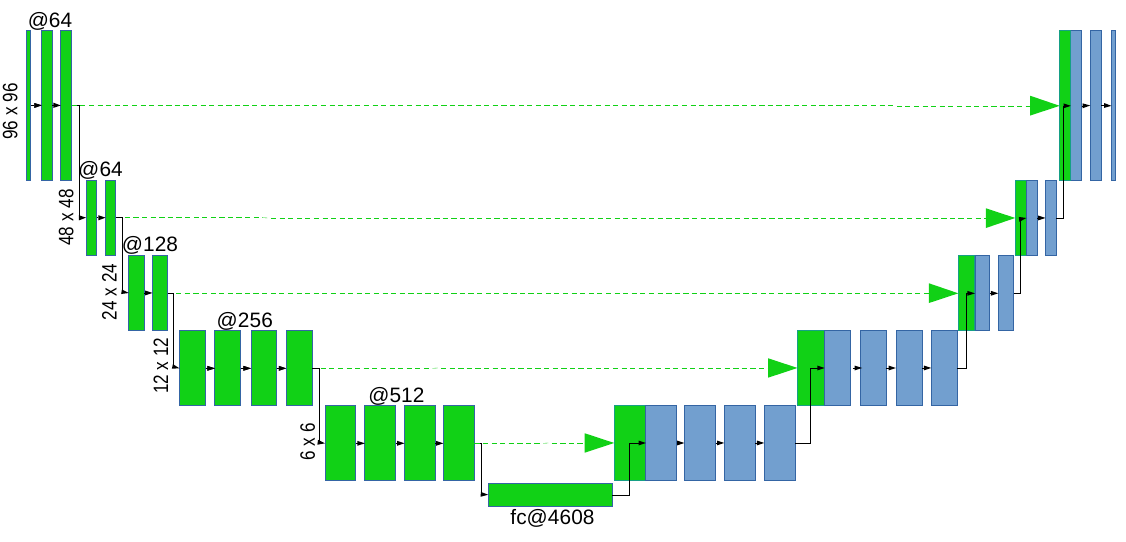}
		\vspace{-10pt}
		\caption{\textbf{Architecture of the MDE generator}. Green: encoder, Blue: decoder. The blocks indicate conv/batch-norm/ReLU layers and the descending/ascending arrows indicate downsampling (strided convolution) and upsampling operators (transposed convolution).}
		\label{generator}
	\end{figure}

	As shown on Figure \ref{generator}, the generator is composed of an encoder and a decoder. The encoder is similar to VGG network, except it only uses one large fully-connected layer at the end. As it is a classical setup in the literature, the decoder mirrors the encoder, but here with the addition of skip-connections to explicitly preserve the non-masked regions.
	
	 The discriminator is very similar to the encoder part of the generator, except that the fully-connected layer is replaced by a global average pooling: as discriminating between real and fake images is considered easier than generating images, it is assumed that the discriminator shall have fewer parameters. As in \cite{radford2015unsupervised}, we use leaky ReLU activations in the discriminator and strided convolutions everywhere instead of max-pooling. We also use a sigmoid layer as the last layer of the generator to better scale the outputs. We use ADAM optimizer with a learning rate of $2.10^{-4}$ for the generator and $2.10^{-5}$ for the discriminator. We train with a momentum of $0.5$ and polynomial learning rate annealing. Finally, we apply $300\,000$ updates with batch size $24$ to train the network.
	
	\section{Experiments}
	
	We validate our method both qualitatively and quantitatively on three datasets, to show its interest for image completion, representation learning as well as face occlusion handling. The \textbf{MNIST} database contains $55\,000$ train and $10\,000$ test images. As MNIST images are grayscale and low resolution, we upscale them to $96 \times 96$ and only apply MDE-RE on this dataset. The \textbf{Oxford-102} flowers dataset consists in 8187 images describing 102 classes of flowers. We train our models on 7167 images from the train and test partitions, and apply them on the 1020 validation images. We report results obtained with MDE-RI, MDE-RE and MDE-REC.
	The \textbf{CelebA} database \cite{liu2015deep} is a large-scale face attribute database which contains $202\,599$ $218 \times 178$ celebrity images coming from $10\,177$ identities, each annotated with $40$ binary attributes (such as \textit{gender}, \textit{eyeglasses}, \textit{smile}), and $5$ landmarks (nose, left and right pupils, mouth corners). As in \cite{zhong2016face}, we use the train partition that contains $162\,770$ images from $8k$ identities to train our models. The test partition contains $19\,962$ instances from $1k$ identities that are different from the training set identities. In all our tests, we use the aligned images, apply a constant rescaling factor ($0.75$) to crop the face region and resize it to $96 \times 96$. All evaluations are performed on the test sets for all datasets.
	
	\subsection{Qualitative evaluation}
	
	\subsubsection{Image completion}
	
	\begin{figure*}[h!]
		\centering
		\includegraphics[width=\linewidth]{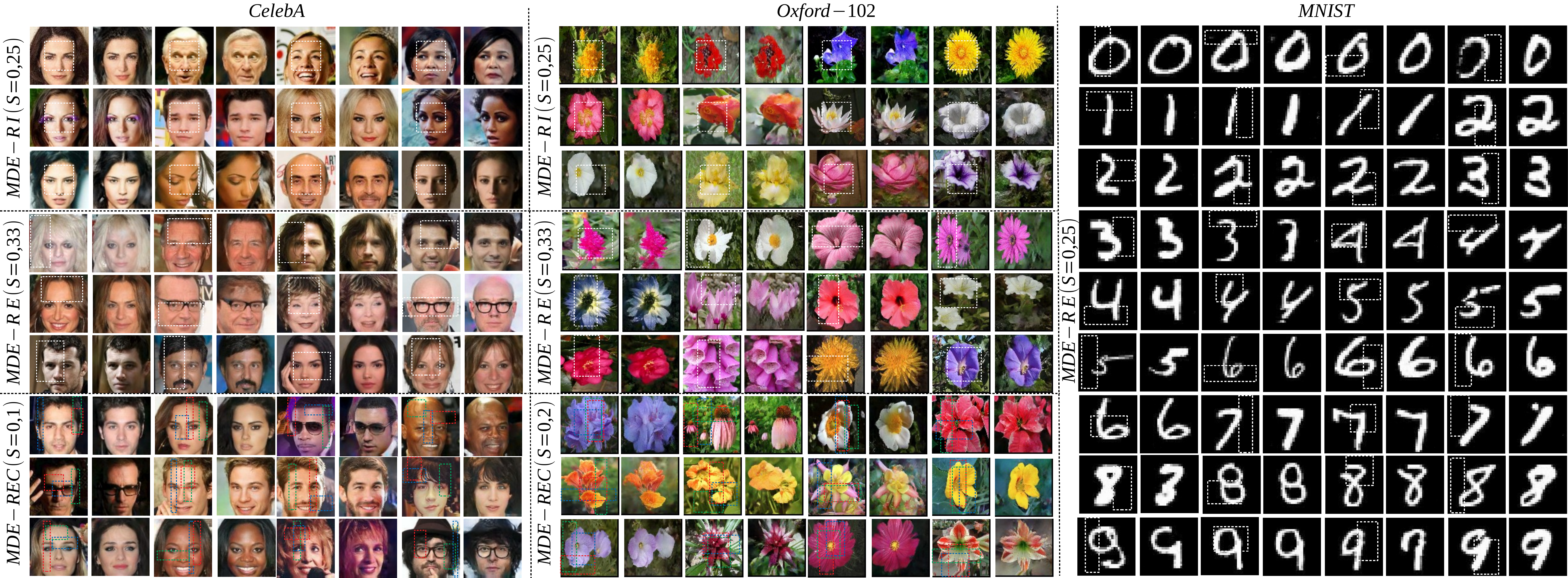}
		\vspace{-10pt}
		\caption{\textbf{Completing images with proposed Missing Data Encoders}. \textit{Left}: Examples of images generated with MDE-RI (3 top rows, $S=0.25$), MDE-RE (3 central rows, $S=0.33$) and MDE-REC (3 bottom rows, $S=0.1$) on CelebA. \textit{Center:} images generated with MDE-RI (3 top rows, $S=0.25$), MDE-RE (3 central rows, $S=0.33$) and MDE-REC (3 bottom rows, $S=0.2$) on Oxford-102. \textit{Right:} Examples of images generated with MDE-RE ($S=0.25$) on MNIST. Images with the dashed boxes are ground truth images and the boxes indicate the non-masked information. For MDE-REC on CelebA and Oxford-102, the red, green and blue boxes show preserved information in R,G,B channels, respectively. Best viewed in color.}
		\label{flowersmde}
	\end{figure*}
	
	\begin{figure*}[h!]
		\centering
		\includegraphics[width=\linewidth]{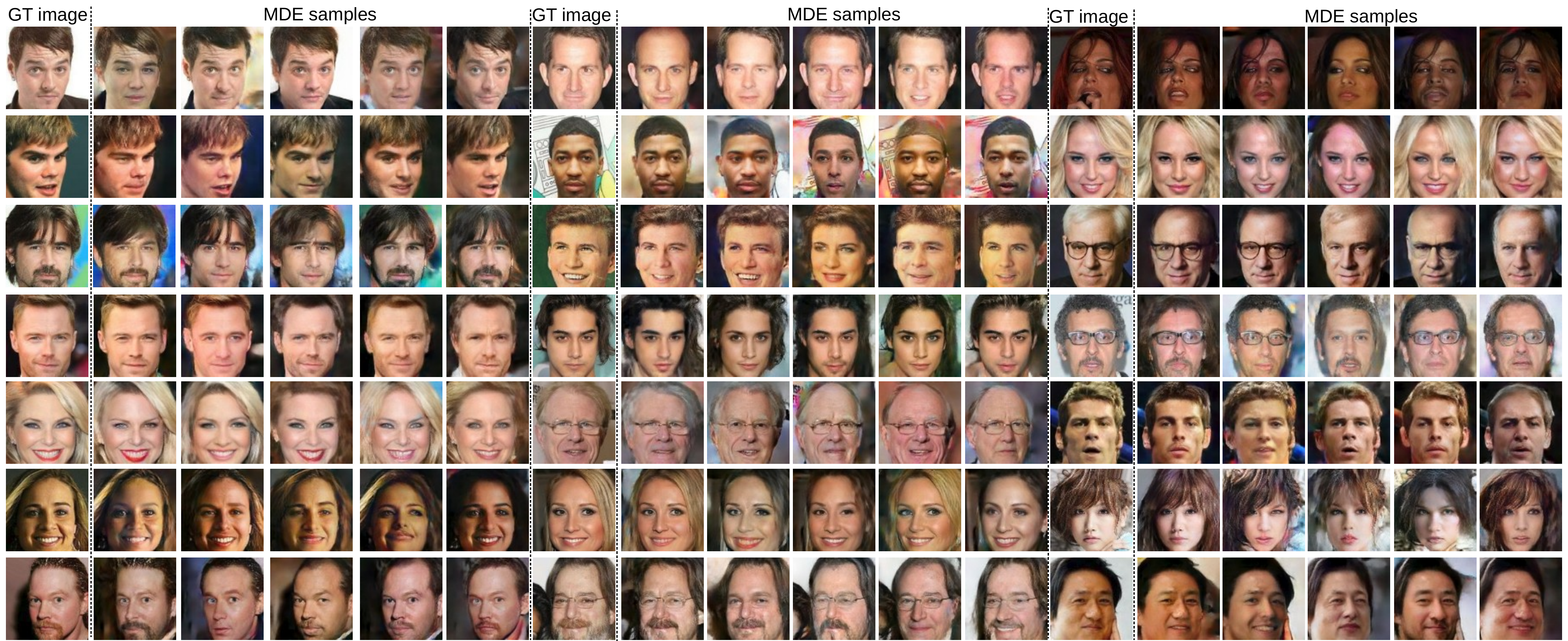}
		\vspace{-10pt}
		\caption{\textbf{Generating multiple image completions with MDE-REC}. For each GT image, a trained MDE (with $S=0.1$) is sampled $5$ times with different input masks.}
		\label{MDE01}
	\end{figure*}

	Figure \ref{flowersmde} shows images generated with MDE on the three datasets. In all cases, the images look very realistic: On MNIST, the generated digits usually match the ground truth ones. On Oxford-102, both the flowers and backgrounds are generated correctly. This implies that even with few data, MDE is able to capture the data distribution.  Similarly, on CelebA, the generated images may present some alterations w.r.t. the ground truth images: the generator may suppress particularly low-probability patterns, such as beards, glasses, hats or a particular facial expression.  Notice however that the quality of the completion is generally high, as there is no blurry pattern or artifact on the generated images. Figure \ref{MDE01} shows more results on CelebA. For each ground truth (GT) image, alternative completions can be generated by applying a new mask before passing the images to the generator. Depending on the mask position and dimensions, the generator can discard background information, or swap haircuts, remove beards or mustache, or change the facial expression.
	
	\subsubsection{MDE resampling}
	
	\begin{figure*}[ht]
		\centering
		\includegraphics[width=0.95\linewidth]{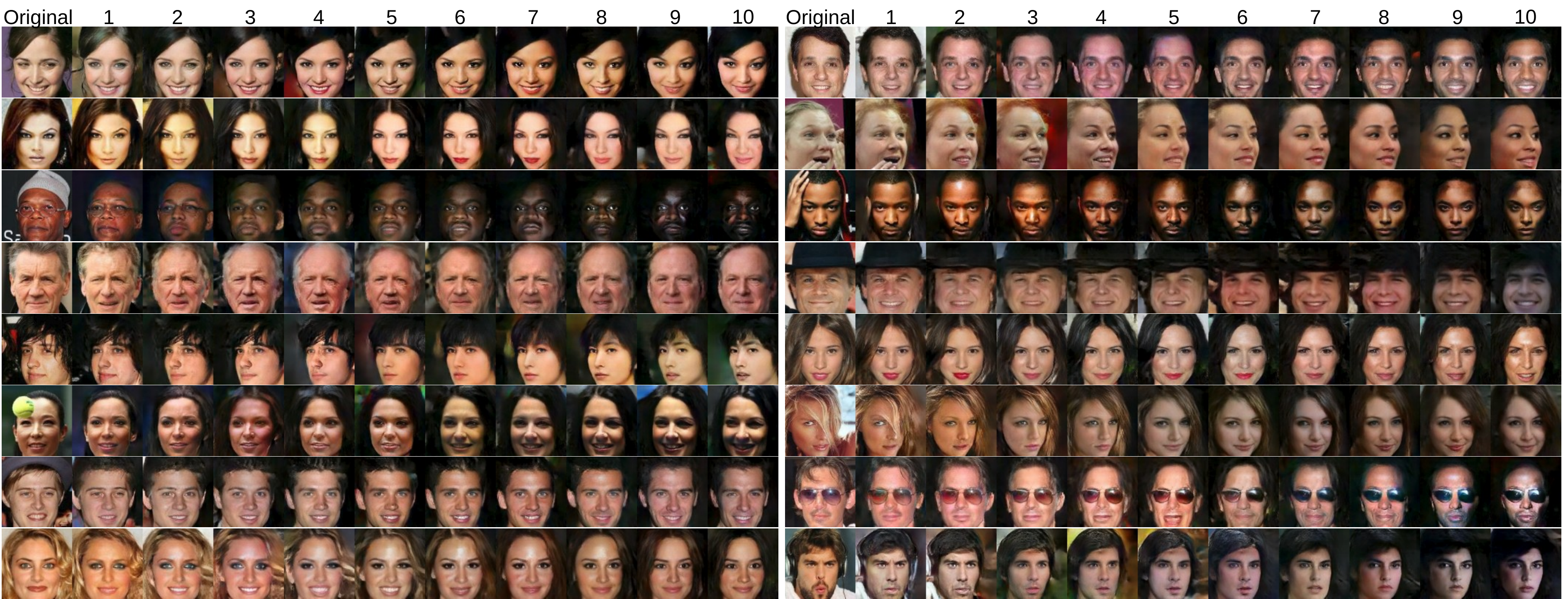}
		\caption{\textbf{Generating sequences of novel images through completion}. Sequences of ten images generated from an original GT one by successive MDE resampling (with $S=0.1$).}
		\label{mderesampling}
	\end{figure*}
	
	Figure \ref{mderesampling} shows sequences of images generated by iteratively resampling an MDE-REC: for a ground truth image, we apply a random mask and generate a new image from it. We then pass the generated image in the generator once again with a new random mask, and repeat these steps $10$ times. Note that at each step, more than $70\%$ of the pixels are completely dropped in average, however the generator generally preserves a lot of semantic information, such as hair color or style, facial expression, gender or ethnicity. After a number of passes through the generator, such information is lost and the faces can be very different from the GT image. The images, however, are still highly realistic, indicating that MDE-REC learns a stable manifold of faces that encompasses face geometry and semantics, which we validate through quantitative evaluations.
	
	\subsection{Quantitative evaluation}
	
	\subsubsection{Evaluation metrics}
	
	We use several metrics to assess the quality of the generated images. The peak signal-to-noise ratio (\textit{pSNR}) quantifies the pixel-wise resemblance between the generated and ground truth images. The structural similarity (\textit{SSIM}) index assesses the holistic visual quality of a completion. Lastly, we measure the \textit{inception score} \cite{salimans2016improved}, which evaluates both the semantic relevance of the generated images as well as their diversity. As computing the inception score requires using a network pretrained on a similar distribution (in our case, a face database), we use VGG-face, as in \cite{wang2018face}. For the same reason, we only perform quantitative evaluation on CelebA.
	
	\subsubsection{Ablation study}
	
	\begin{figure*}[ht]
		\centering
		\includegraphics[width=\linewidth]{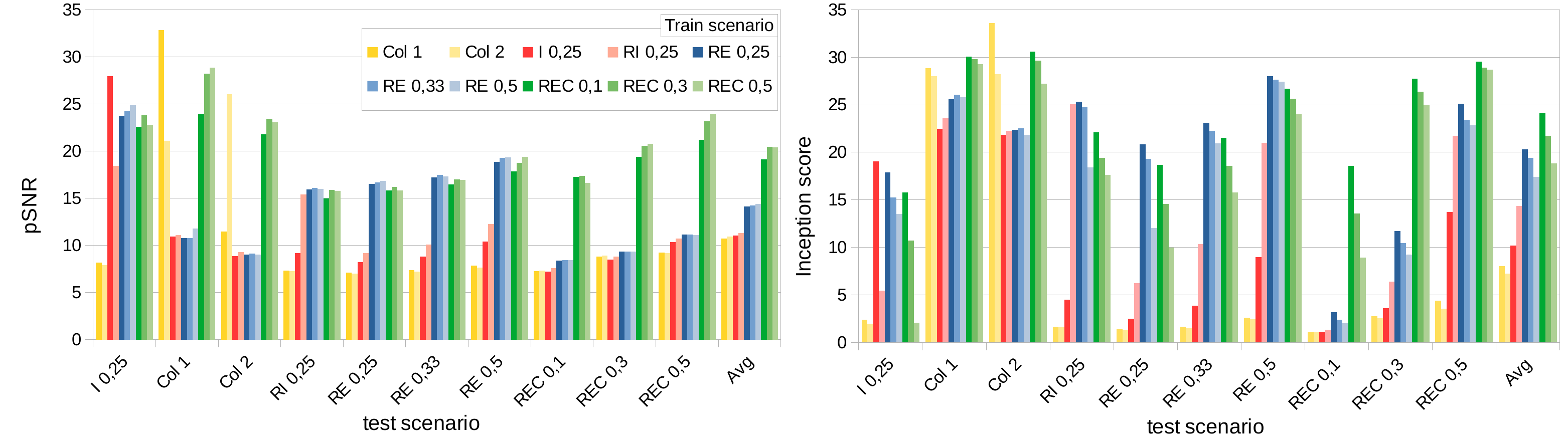}
		\vspace{-13pt}
		\caption{\textbf{Comparing MDE variants in different test setups}. pSNR and Inception score charts for models trained with various tasks and evaluated in different scenarios. I: inpainting. RI: reverse inpainting. Col 1-2: colorization (1-2 channels). RE: random extrapolation. REC: random extrapolation and colorization. For I, RI, RE and REC, appended number indicates the value of masking ratio $S$.}
		\label{MDEsnrsharp}
	\end{figure*}

	\begin{figure}[ht]
		\centering
		\includegraphics[width=\linewidth]{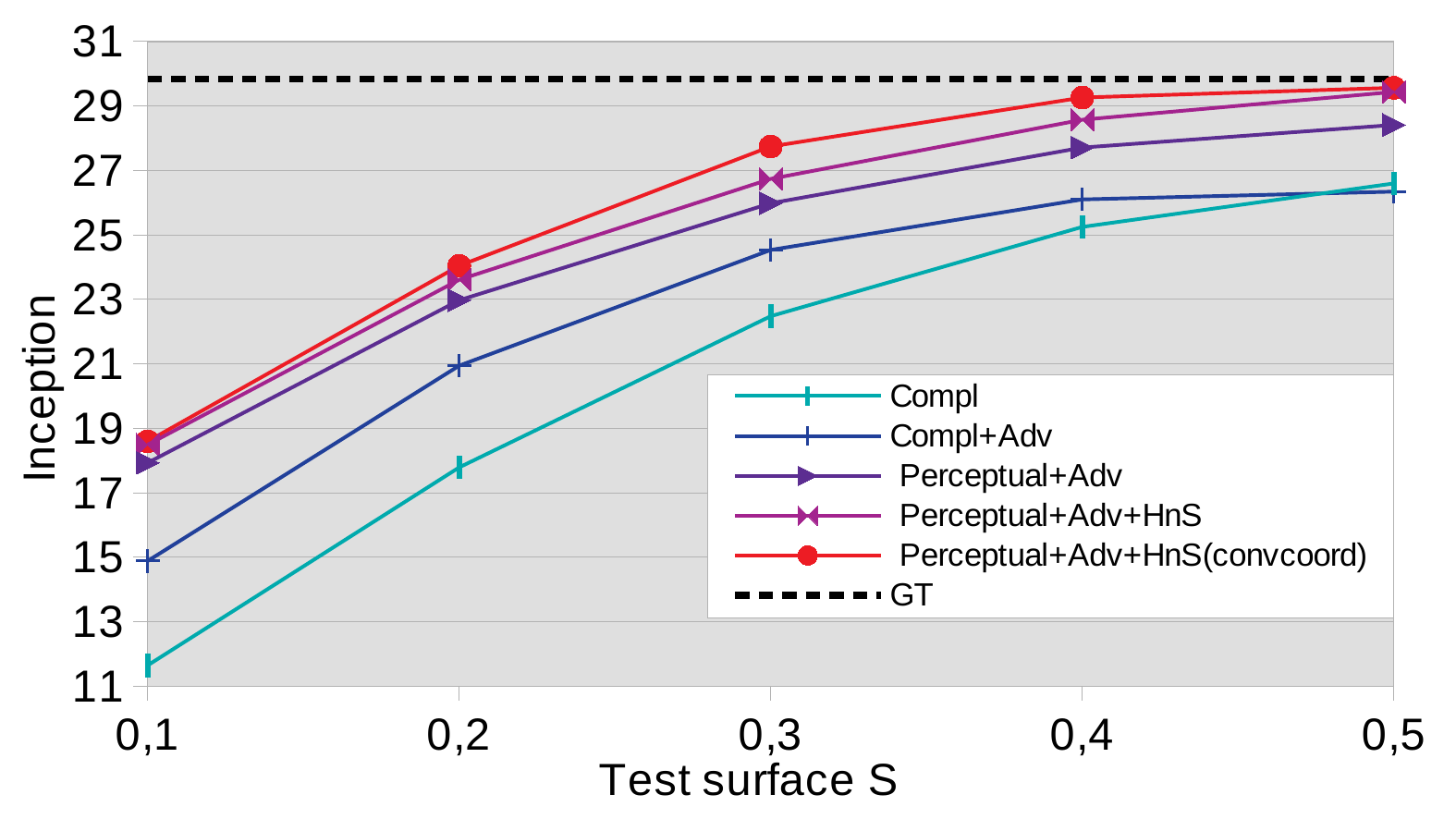}
		\vspace{-13pt}
		\caption{\textbf{Loss ablation}. Inception score for MDE-REC ($S=0.1$) with different loss combinations.}
		\label{MDEsnrsharp2}
	\end{figure}
	
	Figure \ref{MDEsnrsharp} shows pSNR and inception score for multiple train and test scenarios. First, we observe that models trained for colorization, inpainting or reverse inpainting have very low pSNR and inception score when tested in a mismatched scenario (\textit{e.g.} training with inpainting and testing for colorization). On the contrary, MDE-RE also performs well for inpainting and reverse inpainting, as these scenarios can be viewed as special cases of random extrapolation. However, MDE-RE generalizes poorly to colorization as well as REC tasks. Conversely, MDE-REC performs very well on every task both in terms of pSNR and inception score. In terms of pSNR, MDE-RE and MDE-REC trained with high $S$ tend to be better when $S$ is also high in test, and vice-versa, for both RE and REC tasks. However, MDE-RE and MDE-REC trained with $S=0.1$ always have higher inception score. Therefore, MDE-REC with $S=0.1$ is a more generic model that has a better transferability to other completion tasks.
	
	Second, we train a MDE-REC with $S=0.1$ and completion loss, adversarial loss, perceptual loss as well as hide-and-seek loss. Furthermore, we always add a reconstruction loss to ensure that the input information is preserved. We set $\lambda_{rec}^{vgg}=2.10^{-5}$, $\lambda_{adv}=10^{-2}$ and $\lambda_{HnS}=10^{-2}$. Figure \ref{MDEsnrsharp2} draws a comparison between those approaches. As it is classical in the GAN literature, optimizing only $\mathcal{L}_{rec}$ leads to high pSNR/SSIM, but results in blurry images, which have a low inception score. Using adversarial training and \textit{a fortiori} perceptual loss leads to better quality. This is because not only pixel-level information is matched between the generated and ground truth images but also higher-order statistics such as edges intensities for lower VGG layers, and more semantically abstract information for downstream layers. Furthermore, using $\mathcal{L}^{vgg}_{rec}$, $\mathcal{L}_{adv}$ and $\mathcal{L}^{HnS}_{coord}$ yields the best results for every $S$.
	
	\subsubsection{What does MDE learn?}
	
	To study the representations learned by different MDE models, we quantify the transferability of the learned features for attribute prediction and landmark alignment. To do so, we truncate the pretrained MDE models after the fully-connected layer, and append two $4000 \rightarrow 40$ and $4000 \rightarrow 10$ fully-connected sigmoid layers to map the attributes and landmark coordinates, respectively. We then minimize a $L_2$-norm to map these outputs to the 40 attributes and 10 landmarks coordinates, respectively. We perform 5000 updates with batch size 16 (\textit{i.e.} less than one epoch). We report in table \ref{attrlm} the average Euclidean distance between the landmarks as well as the average trace of the confusion matrices.
	
	\begin{table}[htb]
		\centering
		\caption{\textbf{Performance comparison for facial landmark localization and attribute recognition}. Comparison after only 5000 updates. ``Landmarks'': average point-to-point error. ``Attributes'': average trace of the confusion matrices obtained for each attribute.}
		\vspace{5pt}
		\begin{tabular}{l|c|c}
		    \hline
			Pretraining & Landmarks & Attributes\\
			\specialrule{.12em}{.05em}{.05em}
		    Random weights&9.753&19.79\\
			\hline
			Colorization (1c) \cite{zhang2017split}&2.358&11.69\\
			Colorization (2c) \cite{zhang2017split}&2.278&10.92\\
			\hline
			Inpainting \cite{pathak2016context}&5.411&16.12\\
			\hline
			MDE-RI&1,496&13,73\\
			\hline
			MDE-RE(0.25)&2.039&12.72\\
			MDE-RE(0.33)&1.719&11.41\\
			MDE-RE(0.5)&1.759&10.39\\
			\hline
			MDE-REC(0.1)&1.509&10.49\\
			MDE-REC(0.3)&\textbf{1.451}&10.30\\
			MDE-REC(0.5)&1.498&\textbf{10.16}\\
			\hline
		\end{tabular}
		\label{attrlm}
	\end{table}
	
	We observe that reverse inpainting as a pretraining transfers more efficiently to landmark localization and attribute prediction, as compared to inpainting. When compared with colorization, it is less accurate on the attribute prediction task, but better at predicting the face geometry. This stems from the fact that models trained with reverse inpainting only see a limited fraction of the input image. Conversely, MDE-RE models obtain high performance for predicting attributes but a slightly lower accuracy in landmark localization. Finally, MDE-REC models are significantly better for both landmark localization and attribute prediction. Through the channel-wise random region dropping and completion, they benefit from both completion and colorization pre-trainings at the same time. By doing so, they learn to encode the face geometry and high-level semantics in a more efficient way. Note that for both MDE-RE and MDE-REC, the models trained with lower $S$ are not necessarily the best at predicting attributes: this is due to fine-grained attributes such as the presence of earrings or lipstick not being successfully embedded within the generator.
	
	Table \ref{compsota} shows a comparison with recent state-of-the-art approaches, and MDE-REC trained with 50\,000 updates. Our method is competitive with recent methods that use bigger architectures \cite{meyerson2017beyond, he2017adaptively} or pre-training involving large annotated dataset (350k face recognition dataset) \cite{zhong2016face}. This shows that MDE-REC learns useful representations in a completely unsupervised fashion.
	
	\begin{table}[htb]
		\centering
		\caption{\textbf{Facial attribute recognition}. Comparison of unsupervised MDE pre-training with state-of-the-art ($\%$ avg. error).}
		\small
		\vspace{5pt}
		\begin{tabular}{l|r}
		    \hline
			Method&attributes\\
			\specialrule{.12em}{.05em}{.05em}
			Supervised pre-training \cite{zhong2016face}&13.4\\
			\hline
			Single-task baseline \cite{he2017adaptively}&10.37\\
			\hline
			Multi-task baseline \cite{he2017adaptively}&9.58\\
			\hline
			Parallel order \cite{meyerson2017beyond}&10.21\\
			\hline
			Parallel order+landmarks \cite{meyerson2017beyond}&10.29\\
			\hline
			Soft order+identity \cite{meyerson2017beyond}&\textbf{8.64}\\
			\hline
			\hline
			MDE-REC(0.5)&9.17\\
			\hline
		\end{tabular}
		\label{compsota}
	\end{table}
	
	\subsubsection{Face completion under targeted occlusions:}
	
	\begin{table}[htb]
		\centering
		\caption{\textbf{Comparison with state-of-the-art for face completion under targeted occlusions}. Results for context encoder (CE \cite{pathak2016context}) and generative face completion (GFC \cite{li2017generative}) are excerpted from \cite{li2017generative}.}
		\small
		\vspace{5pt}
		\begin{tabular}{c|c|c|c|c|c|c}
		    \hline
			&\multicolumn{3}{c|}{pSNR}&\multicolumn{3}{c}{SSIM}\\
			\specialrule{.12em}{.05em}{.05em}	
			Occlusion&CE &GFC&MDE&CE&GFC&MDE\\
			\hline
			Right half&18.6&19.4&\textbf{21.6}&0.772&0.804&\textbf{0.814}\\
			\hline
			Left half&18.4&19.3&\textbf{21.8}&0.774&0.808&\textbf{0.815}\\
			\hline
			Both eyes&17.9&18.3&\textbf{21.8}&0.719&0.731&\textbf{0.839}\\
			\hline
			Right eye&19.0&19.1&\textbf{22.4}&0.754&0.759&\textbf{0.855}\\
			\hline
			Left eye&19.1&18.9&\textbf{22.6}&0.757&0.755&\textbf{0.860}\\
			\hline
			Mouth&19.3&19.7&\textbf{21.9}&0.818&\textbf{0.824}&0.818\\
			\hline
		\end{tabular}
		\label{targetedocclusions}
	\end{table}
	
	We also study the application of MDE-REC ($S=0.1$) to face completion under occlusions. We use the same protocol as in \cite{li2017generative} and compare with state-of-the-art methods, CE \cite{pathak2016context} and GFC \cite{li2017generative}, without postprocessing. The results show that MDE-REC is more efficient than the random inpainting proposed by \cite{li2017generative}. In addition, our method is agnostic to the nature of the dataset, as opposed to \cite{li2017generative}, where the authors use an auxiliary face parsing network. As shown in Table \ref{targetedocclusions}, results for MDE are significantly better nearly everywhere. Furthermore, high values of the inception score (which ranges from 18.80 to 27.28) indicates that the generated images are sharp and realistic.
	
	\section{Conclusion}
	
    In this paper, we introduced the Missing Data Encoder for image completion, unsupervised representation learning and face occlusion handling. The network is trained to complete an image from a rectangular region drawn at random in each channel independently, a task that subsumes to some extent inpainting, reverse inpainting and colorization.
    We showed on several datasets that the proposed method allows the generation of high quality images from only small seed fragments. By learning to do so, our architecture captures without supervision high level semantic information within its embedding. It also extends the state-of-the-art for face completion under occlusion. Future work involves using MDE pretraining for classification or semantic segmentation, as well as investigating the use of the proposed ``hide and seek'' adversarial loss for other applications such as object detection.

{\small
	\bibliographystyle{ieee}

}

\end{document}